\documentclass[journal]{IEEEtran}
\usepackage[noend]{algpseudocode}
\usepackage{algorithmicx,algorithm}
\usepackage{graphicx}
\usepackage[colorlinks,linkcolor=black,citecolor=black]{hyperref}
\usepackage{amssymb}
\usepackage{booktabs}
\usepackage{amssymb}
\usepackage{indentfirst}
\usepackage[misc]{ifsym}
\usepackage{amsmath}
\usepackage{multirow}
\usepackage{booktabs}
\usepackage{tikz,xcolor,hyperref}
\usepackage{bbding}
\usepackage{underscore}
\usepackage[T1]{fontenc}
\usepackage{amssymb}
\usepackage{amssymb}
\usepackage[misc]{ifsym} 
\usepackage{float}
\usepackage{amssymb}
\usepackage{amssymb}
\usepackage{color}
\usepackage{amssymb}
\usepackage{indentfirst}
\usepackage[misc]{ifsym}
\usepackage{booktabs}
\usepackage{tikz,xcolor,hyperref}
\usepackage{bbding}
\usepackage{array}
\usepackage{colortbl}
\usepackage{bbding}
\usepackage{pifont}
\usepackage{wasysym}
\usepackage{amssymb}
\usepackage{cite}

\makeatletter

\newcommand{\Rmnum}[1]{\expandafter\@slowromancap\romannumeral #1@}
\makeatother
\usepackage{lineno} 
\begin{document}

%
\title{Temporal Prototyping and Hierarchical Alignment for Unsupervised   Video-based Visible-Infrared Person Re-Identification}
%
%
%

\author{ Zhiyong Li, Wei Jiang, Haojie Liu, Mingyu Wang, Wanchong Xu, Weijie Mao

	\thanks{This work was supported by the Zhejiang Province Natural Science Foundation of China (Grant LZ24F030004). \textit{(Corresponding author: Wei Jiang, e-mail: jiangwei\_zju@zju.edu.cn)}}
	\thanks{ Zhiyong Li,Haojie Liu, Mingyu Wang, Wanchong Xu and Weijie Mao are with the College of Control Science and Engineering, Zhejiang University, Hangzhou 310027, China. (email: lizhiyong\_zju@zju.edu.cn, liuhaojie@zju.edu.cn, wangmingyu@zju.edu.cn,  3210104804@zju.edu.cn, wjmao@zju.edu.cn )}
	\thanks{Wei Jiang is with the School of Computer Science and Technology, Zhejiang University of Water Resources and Electric Power, Hangzhou, 310018, China (e-mail: jiangwei\_zju@zju.edu.cn).}
	
}

\maketitle

\begin{abstract}
	Visible-infrared person re-identification (VI-ReID) enables cross-modality identity matching for all-day surveillance, yet existing methods predominantly focus on the image level or rely heavily on costly identity annotations. While video-based VI-ReID has recently emerged to exploit temporal dynamics for improved robustness, existing studies remain limited to supervised settings. Crucially, the unsupervised video VI-ReID problem, where models must learn from RGB and infrared tracklets without identity labels, remains largely unexplored despite its practical importance in real-world deployment. To bridge this gap, we propose \textbf{HiTPro} (\textbf{H}ierarchical \textbf{T}emporal \textbf{Pro}totyping), a prototype-driven framework without explicit hard pseudo-label assignment  for unsupervised video-based VI-ReID. HiTPro begins with an efficient Temporal-aware Feature Encoder that first extracts discriminative frame-level features and then aggregates them into a robust tracklet-level representation. Building upon these features, HiTPro  first constructs reliable intra-camera prototypes via {Intra-Camera Tracklet Prototyping}  by aggregating features from temporally partitioned sub-tracklets. Through {Hierarchical Cross-Prototype Alignment} , we perform a two-stage positive mining process: progressing from within-modality associations to cross-modality matching, enhanced by Dynamic Threshold Strategy  and Soft Weight Assignment. Finally, {Hierarchical Contrastive Learning}  progressively optimizes feature-prototype alignment across three levels: intra-camera discrimination, cross-camera same-modality consistency, and cross-modality invariance. Extensive experiments on HITSZ-VCM and BUPTCampus demonstrate  that HiTPro achieves state-of-the-art performance under fully unsupervised settings, significantly outperforming adapted baselines and establishes  a strong baseline for future research. Code will be available at https://github.com/ThomasjonLi/HiTPro. 
	 
\end{abstract}

\section{Introduction}
Visible-infrared person re-identification (VI-ReID) aims to match pedestrian identities across non-overlapping cameras operating in heterogeneous spectral bands, typically visible (RGB) and infrared (IR), enabling robust 24/7 surveillance under varying illumination conditions\cite{huang2022deep,ye2021deep,leng2019survey}. Since the introduction of the first large-scale benchmark, SYSU-MM01\cite{Wu2017}, VI-ReID has drawn significant attention due to its practical value and the inherent challenge of bridging the substantial cross-modality discrepancy. 
To address this modality gap,  VI-ReID methods predominantly operated on static images, focusing on learning modality-invariant representations through shared embedding spaces \cite{Feng2019,lu2023learning,Lu2020} or cross-modality images synthesis methods \cite{zhong2021grayscale,liu2022cross,qi2023generative}.

However, real-world surveillance systems typically capture video sequences, which contain rich temporal dynamics such as gait, motion patterns, and pose transitions that are invaluable for robust identity matching, especially under occlusion or low-quality imaging conditions. This has motivated the emergence of video-based Re-ID. In the single-modality (visible) setting, 
researchers have adopted attention mechanisms or Transformer-based architectures \cite{liu2017video, chang2018multi,lin2022learning}. Recently, several works have  extended video Re-ID to the visible-infrared domain. For instance, Lin et al. \cite{lin2022learning} proposed a temporal memory module on the HITSZ-VCM dataset to align modality-shared motion patterns, while Du et al. \cite{du2023video} introduced auxiliary visible samples to enhance infrared representation learning on BUPTCampus. Despite this progress, a critical limitation persists: all existing video VI-ReID approaches assume full identity supervision during training. In real-world scenarios, however, collecting large-scale labeled cross-modal video datasets is prohibitively expensive and often infeasible due to privacy constraints and camera heterogeneity. This necessitates the study of unsupervised video VI-ReID, where models must learn transferable representations without any identity annotations.

The shift from supervised to unsupervised video-based visible-infrared person re-identification introduces fundamental challenges that cannot be adequately addressed by simply extending existing image-level paradigms. Most current unsupervised VI-ReID methods \cite{shi2025multi,yang2023dual,yang2022augmented} rely on a two-stage pipeline: first performing intra-modality clustering to generate pseudo-labels, and then aligning these clusters across modalities via contrastive learning or graph matching. While effective for static images, this strategy breaks down in the video setting. 
First, the number of tracklets is drastically smaller than that of individual frames, which violates the density assumption of clustering algorithms like DBSCAN \cite{ester1996density}. 
Second, tracklet representations exhibit large intra-identity variation due to pose dynamics, motion blur, and background clutter across the sequence. This leads to less compact feature distributions, making it difficult for clustering to separate identities reliably.  
Third, and  uniquely to VI-ReID, each identity must be matched across two orthogonal sources of heterogeneity: cross-camera  shifts, and cross-modality gaps. However, conventional unsupervised pipelines, originally designed for either single-modality ReID or image-level VI-ReID, lack a mechanism to  simultaneously address the aforementioned issues.

\begin{figure}[ht]
	\centering
	\includegraphics[width=1\linewidth]{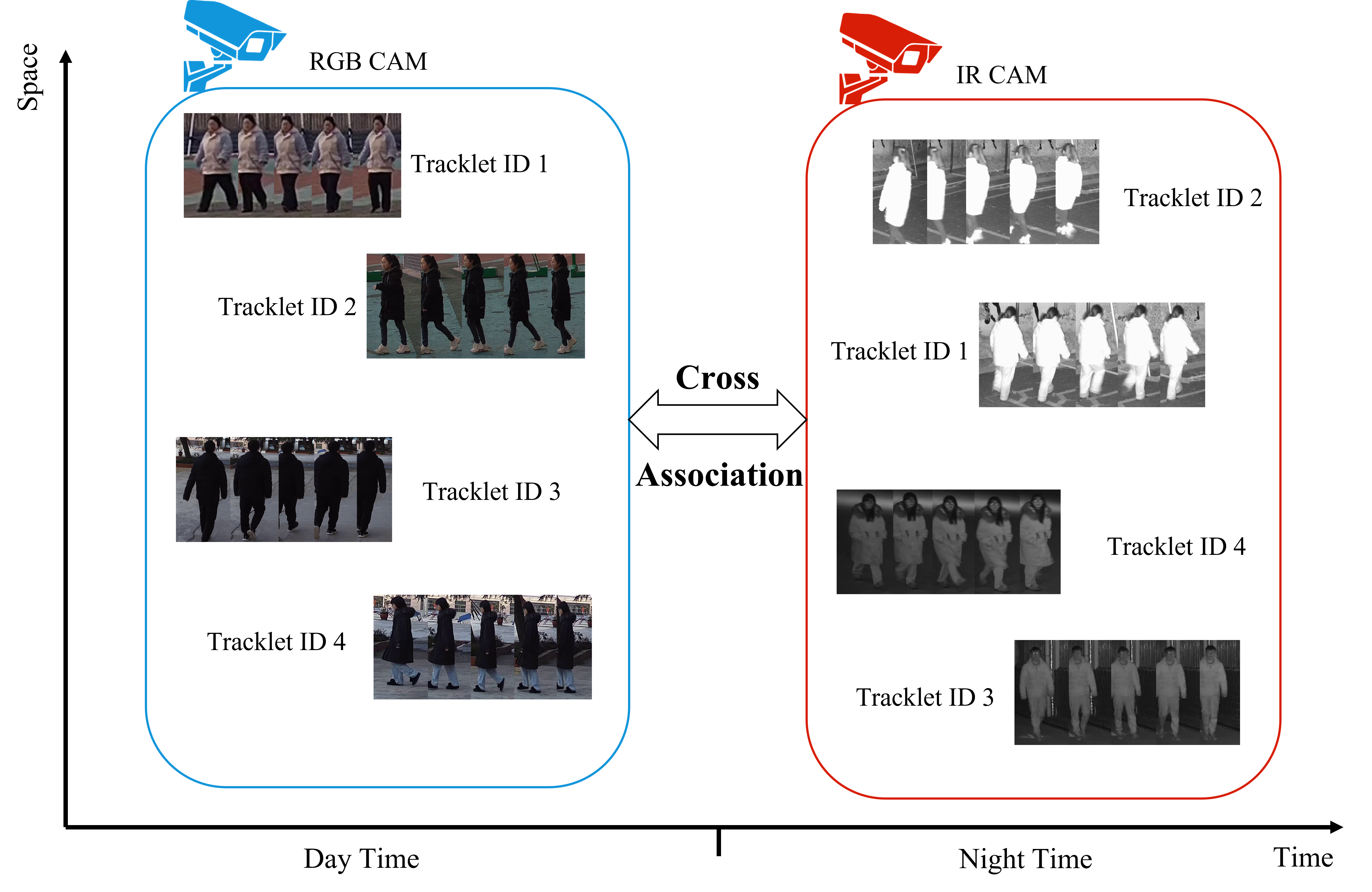}
	\caption{Illustration of spatio-temporal variations in unsupervised video-based visible-infrared person re-identification: within a single camera over a short time window, multiple pedestrians occupy distinct spatial regions, implying that their tracklets must correspond to different identities; Conversely, the same individual may appear across different cameras at temporally disjoint intervals, typically as a visible tracklet during daytime and an infrared tracklet at night.
	}
	\label{fig:illustration}
\end{figure}

To this end, we revisit cross-modal video person re-identification from the perspective of real-world surveillance systems.  Unlike image-based person ReID, pedestrians in video surveillance inherently exhibit spatio-temporal variations \cite{li2019unsupervised}. As illustrated in  Figure \ref{fig:illustration}, within a single camera over a short time window, multiple pedestrians occupy distinct spatial regions, implying that their tracklets must correspond to different identities; this intra-camera spatial separation provides a strong structural prior for identity discrimination. Conversely, the same individual may appear across different cameras at temporally disjoint intervals, typically as a visible tracklet during daytime and an infrared tracklet at night, requiring unsupervised association across both camera views and  modalities despite the absence of overlapping observations or labels.

Motivated by these insights, we propose HiTPro (Hierarchical Temporal Prototyping), an unsupervised framework that leverages spatio-temporal structure in surveillance video for prototype-based learning. HiTPro begins with an efficient Temporal-aware Feature Encoder (TFE) that first extracts frame-level features for each video frame, and then aggregates them within each tracklet via temporal attention, suppressing transient noise and producing a stable tracklet-level representation. Built upon this foundation, HiTPro introduces two core components. First, to address sample sparsity and avoid unreliable clustering, we design Intra-Camera Tracklet Prototyping (ICTP), which leverages the structural prior that all tracklets observed by a single camera in  a  short period time must correspond to distinct identities. Rather than treating each tracklet as a monolithic unit, ICTP  partitions it into multiple sub-tracklets along the temporal axis. Each sub-tracklet is encoded independently, and their features are aggregated, to form a  prototype per tracklet. 
Second, to jointly resolve cross-camera appearance shifts and cross-modality gaps, we propose Hierarchical Cross-Prototype Alignment (HCPA). Instead of direct feature alignment, HCPA performs a two-stage positive mining process grounded in prototype similarity. It begins with intra-modality cross-camera mining: for each tracklet prototype, high-confidence matches are identified across other cameras within the same modality. Then we perform cross-modality mining, matching visible prototypes to infrared ones (and vice versa). To ensure reliability throughout training, HCPA employs a Dynamic Threshold Strategy (DTS) that adaptively adjusts the similarity threshold per instance based on its current matching confidence, preventing premature exclusion of valid positives. 
The resulting positive pairs are further refined by Soft Weight Assignment (SWA), which assigns soft weights according to relative similarity strength to emphasize more reliable matches during alignment. 

Finally, building on the reliable positive pairs mined by HCPA, we introduce Hierarchical Contrastive Learning (HCL) that explicitly optimizes the alignment between sub-tracklet features and their corresponding identity prototypes. This learning is organized hierarchically: it first enforces intra-camera separation by repelling sub-tracklets from different identities within the same camera; then aligns sub-tracklets to cross-camera prototypes of the same identity within the same modality; finally, it bridges the visible-infrared gap by attracting sub-tracklets to cross-modality prototypes weighted by SWA confidence. 

HCL ensures that fine-grained temporal fragments consistently converge to a unified, modality-invariant identity anchor. 
In summary, our main contributions are listed as follows:

$\bullet$ We introduce the problem of unsupervised video-based visible-infrared person re-identification, highlighting its unique challenges beyond image-level or single-modality settings. To address this problem, we propose HiTPro, a dedicated framework built upon a Temporal-aware Feature Encoder (TFE) to aggregate robust tracklet representations.

$\bullet$ We introduce Intra-Camera Tracklet Prototyping (ICTP) and Hierarchical Cross-Prototype Alignment (HCPA), two complementary strategies that exploit camera-wise identity exclusivity and trajectory-level cross-modal consistency to mine high-confidence positive pairs without supervision.

$\bullet$ We design Hierarchical Contrastive Learning (HCL), which directly aligns sub-tracklet features with identity prototypes in a progressive manner: from intra-camera discrimination, to cross-camera same-modality alignment, and ultimately to cross-modality matching.

$\bullet$ We conduct comprehensive experiments on existing video-based VI-ReID datasets and demonstrate that HiTPro achieves state-of-the-art performance under  unsupervised settings, validating the effectiveness of our proposed learning paradigm.

\section{Related Works}
\subsection{Image-based Visible-Infrared Person Re-Identification}
Visible-infrared person re-identification (VI-ReID) has emerged as a critical task for all-day surveillance, aiming to match pedestrian identities across visible  and infrared  imaging modalities under large cross-modality appearance discrepancies. Early supervised approaches primarily focus on mitigating modality gaps through shared embedding learning, particularly by the design of the two-stream convolutional neural network \cite{Wu2017,Ye2018a,Ye2020,ye2021deep}. 
Recent works have also designed two-stream architectures based on the emerging Transformer framework \cite{chen2022structure,chai2023dual,jiang2022cross}.
Beyond structural alignment, a second line of research leverages image-level generative models such as GAN to translate images across spectra, thereby enabling inter-modality matching \cite{Liu2023,zhong2021grayscale,liu2022cross,qi2023generative}. On this basis , feature-level generation or transformation has gained traction as a more stable alternative that avoids the instability of image synthesis \cite{zhang2022fmcnet,zhang2023diverse,liu2025stochastic}. These methods operate in latent space to either hallucinate modality-invariant features or explicitly disentangle identity-related and modality-specific component. Recently, cross-modal learning based on frequency-domain feature analysis and separation has also begun to attract researchers' attention \cite{zhang2025frequency,gu2025discovering,yu2026wavelet}.

However, the reliance on expensive cross-modal identity annotations severely limits real-world deployment. This has spurred growing interest in unsupervised VI-ReID. H2H \cite{liang2021homogeneous} first  introduce this problem and proposed a two-stage learning strategy that progresses from intra-modality clustering-based learning to cross-modality alignment. OTLA \cite{wang2022optimal} proposes a method based on optimal transport paths to transfer labels from the visible  to  infrared modality. Subsequent methods have focused on improving clustering strategies and cross-modal alignment techniques \cite{wu2023unsupervised,pang2023cross,li2024inter,yang2024shallow,yin2025adaptive}. Although image-level unsupervised VIReID methods have achieved certain success on existing datasets, they operate at the image level, ignoring the rich temporal dynamics in video surveillance sequences, such as pose and motion changes, as well as environmental disturbances, that are crucial for robust identity matching. This limitation prevents them from learning stable identity-aware representations when applied to video-level features.

\subsection{Video-based Person Re-Identification}
Video-based person re-identification exploits temporal information from tracklets to enhance robustness against occlusion, pose variation, background clutter, and transient appearance changes. Early approaches adopt recurrent networks\cite{mclaughlin2016recurrent,zhou2017see} or 3D-convolutions \cite{gu2020appearance,li2019multi} to capture motion cues across frames. 
Later works leverage attention mechanisms to weight informative frames \cite{wang2021pyramid,chen2020temporal}. Recent advances focus on fine-grained temporal structure, long-range dependency modeling, and explicit preservation of spatial identity cues across time \cite{hou2020temporal,yang2020spatial}. 
Complementing these sequence modeling efforts, spatial-aware or fine-grained based  frameworks have been developed to preserve spatial semantics during temporal fusion \cite{fu2019sta,ran2024multiscale,liu2019spatial,ma2025recursively}. 
                                      
Some researchers have also proposed unsupervised learning methods  designed for single-modality video ReID.  TAUDL \cite{li2018unsupervised} and UTAL \cite{li2019unsupervised} associate tracklets across cameras using spatio-temporal coherence and appearance consistency without any identity annotations. 
Several research  also focus on  video feature clustering method optimization \cite{xie2021unsupervised,qian2022successive,xie2022sampling,zhang2025dual}.  
To address feature diversity in unlabeled videos, Zhang et al. \cite{zhang2025dual} introduce a dual-modeling approach combining reliable tracklet representations from easy frames and informative representations from hard frames, enhanced by discrepant memory mining. 
However, these methods are developed exclusively in  visible-only settings. They inherently assume relatively small appearance variations across views, and thus fail catastrophically in cross-modality scenarios like RGB-IR matching, where modality discrepancy dominates over identity signals. Consequently, none can be directly applied to the unsupervised video-based VI-ReID problem addressed in this work.

\subsection{Video-based Visible-Infrared Person Re-Identification}
A few recent works have begun to investigate supervised video-based VI-ReID by extending temporal modeling to cross-modality settings. Lin et al. \cite{lin2022learning} first introduce video-based learning into VI-ReID and contribute  a large scale video-based cross-modal ReID dataset HITSZ-VCM. They
propose a temporal-memory to  captures discriminative motion patterns. 
Li et al. \cite{li2023intermediary} employ bidirectional LSTM to build temporal information and  leverages anaglyph data as cross-modal bridges. 
FA-Net \cite{yang2025fa} aligns spatial information with global-local feature fusion and achieves temporal alignment by  learnable weighting. SIMFGA \cite{zuo2025spatio} also integrates local features with a temporal Transformer to jointly align spatio-temporal information across dual modalities. STHF \cite{tao2026spatial} proposed a high-pass filter to suppress spatial-temporal low-frequency components in tracklets.
Current methods mostly focus on jointly aligning spatio-temporal information through network design. However, all these approaches strictly rely on full identity supervision and offer no mechanism to handle the unlabeled setting, making them inapplicable to real-world deployments where cross-modal tracklet annotations are scarce or absent. In this work, we address the problem of unsupervised video-based VI-ReID and propose a novel HiTPro framework. The proposed method leverages frame-level consistency within unlabeled tracklets to construct reliable pseudo-prototypes, while enforcing cross-camera and cross-modality alignment, enabling robust and scalable identity association without labeled video data.

\section{Method}
\begin{figure*}[ht]
	\centering
	\includegraphics[width=1\linewidth]{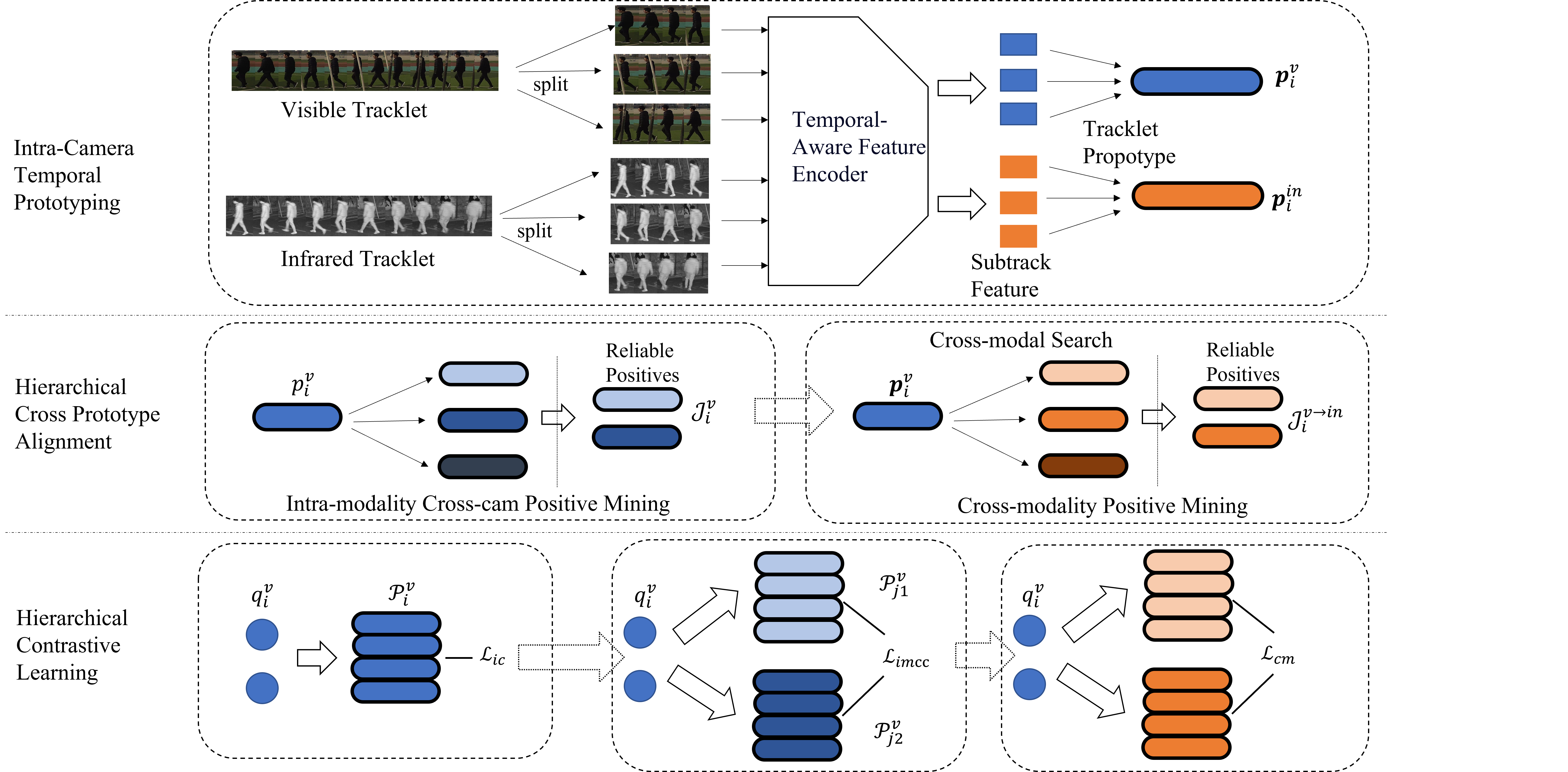}
	\caption{Illustration of the proposed HiTPro framework. The input video tracklets from visible and infrared modalities are processed by a  Temporal-aware Feature Encoder (TFE) to obtain modality-invariant temporal-aggregated features. Each tracklet is split into sub-tracklets, whose embeddings are aggregated into tracklet-level prototypes. The HCPA module then performs hierarchical positive mining: first within modality across cameras, then across modalities, using dynamic thresholds and soft weights to ensure reliable matching. During training, a curriculum-based contrastive loss progressively aligns representations at intra-camera, intra-modality, and cross-modality levels.}
	\label{fig:structure}
\end{figure*}

In this section, we present HiTPro, a Hierarchical Temporal Prototyping framework for unsupervised cross-modal video person re-identification. An overview of the HiTPro architecture is shown in Fig. \ref{fig:structure}. Section \ref{TFE}  introduces the Temporal Feature Encoder (TFE), which enhances video-level representations by capturing inter-frame dynamics and adaptively emphasizing discriminative frames. Building upon these features, Section \ref{ICTP} describes Intra-Camera Temporal Prototyping (ICTP)—a self-supervised strategy that constructs tracklet prototypes within each camera by exploiting the identity-disjointness prior of pedestrian tracks. To bridge heterogeneous views, Section \ref{HCPA} details our Hierarchical Cross Prototype Alignment mechanism, which progressively identifies reliable cross-camera and cross-modal positive prototypes through similarity-guided soft labeling. Finally, Section \ref{HCL} presents the hierarchical contrastive learning framework, which progressively optimizes intra-camera, intra-modality cross-camera, and cross-modality alignment through a curriculum-based loss scheduling strategy.

\subsection{Temporal-aware Feature Encoder (TFE).} \label{TFE}
\begin{figure}[ht]
	\centering
	\includegraphics[width=1\linewidth]{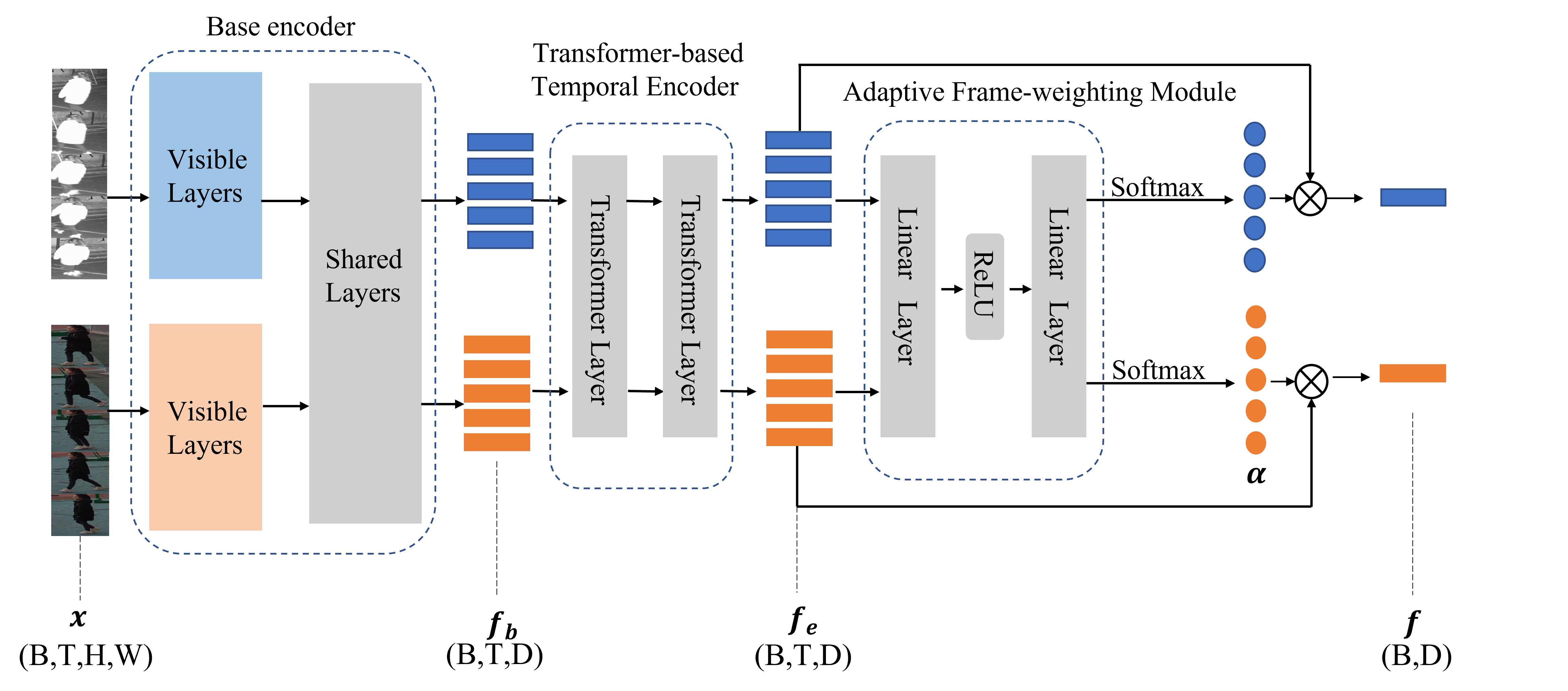}
	\caption{Architecture of the Temporal-aware Feature Encoder (TFE). Input video clips from visible and infrared  modalities are first processed by the base encoder. Subsequently, a two-layer Transformer encoder models long-range temporal dependencies across frames. Finally, an adaptive frame-weighting module computes attention scores for each frame and produces a compact, discriminative video-level embedding by weighted aggregation.}
	\label{fig:TFE}
\end{figure}

To effectively model the temporal dynamics of video sequences and extract discriminative person representations under cross-modal settings, we design a Temporal-aware Feature Encoder (TFE) that integrates a dual-stream backbone with  temporal aggregation modules. 
As illustrated in Figure \ref{fig:TFE}, TFA first processes input video clips $\mathbf{x} \in \mathbb{R}^{T \times H \times W}$ from visible and infrared modalities through a shared backbone network derived from the AGW framework \cite{ye2021deep}. Specifically, each frame in the video snippet is processed through a unified pipeline: it first passes through modality-specific shallow layers to capture surface-level semantic cues, and then enters a shared deep backbone to extract modality-invariant identity representations $\mathbf{f_b}$. Building upon these frame features, the Transformer-based Temporal Encoder (TTE), which consist of two Transformer Layer,  is employed to model long-range dependencies between frames by capturing global contextual information via self-attention mechanisms. This enables the network to understand complex motion patterns and occlusion scenarios, thereby enhancing the robustness of individual frame representations.
The enhanced features $\mathbf{f_e}$ are then fed into an Adaptive Frame-weighting Module (AFM), which computes importance scores for each frame using an attention network composed of two linear layers with ReLU activation and a softmax layer. Formally, given a sequence of enhanced features $\mathbf{f_e}=\{f_{e_1},f_{e_2},...,f_{e_T}\}\in \mathbb{R}^{B \times T \times D}$ The attention weights are computed as:
\begin{equation}
	\alpha_t = \text{Softmax}\left( \mathbf{w}^\top \phi({f}_{e_t}) \right)
\end{equation}
where $\phi(\cdot)$ is a learnable projection function and $w$ is a weight vector. The final video-level embedding is obtained by weighted summation:
\begin{equation}
	f = \sum_{t=1}^{T} \alpha_t\cdot f_{e_t}
\end{equation}
This design allows TFE to dynamically emphasize informative frames (e.g., clear appearance or full body) while suppressing noisy or occluded ones, resulting in a compact and discriminative representation suitable for downstream contrastive learning.

\subsection{Intra-Camera Temporal Prototyping (ICTP)} \label{ICTP}
To enable self-supervised learning in the absence of identity labels, we propose Intra-Camera Temporal Prototyping (ICTP), which constructs trajectory-level prototypes by leveraging a natural structural prior in video surveillance: within a single camera, all observed pedestrian tracklets are mutually identity-disjoint, i.e., no two distinct tracklets correspond to the same person at the same time. This property guarantees that each tracklet represents a unique underlying identity, providing a reliable basis for prototype formation without cross-camera association or clustering.

Given a set of tracklets $X^{m,c}=\{x_1,x_2,...,x_{N_{c}}\}$ from modality m (visible or infrared), where $N_{c}$ denotes the number of tracklets in camera c, we first partition each long tracklet $x_i$ into K non-overlapping sub-tracklets $\mathcal{I}_i=\{x_i^{1},x_i^{2},...,x_i^{K}\}$. This subdivision captures local appearance variations (e.g., due to pose change or partial occlusion) while preserving the identity consistency within the original tracklet. Each sub-tracklet $x_i^k$ is processed by the Temporal Feature Encoder to produce a compact video-level embedding:
\begin{equation}\label{eq:fTFE}
	\mathbf{f}_{i}^k = \text{TFE}\left( x_i^{k} \right) \in \mathbb{R}^D
\end{equation}
The trajectory prototype $p_i$ for tracklet $x_i$ is then defined as the mean feature of all its sub-tracklet features:

\begin{equation}\label{eq:constructp}
	{p}_{i} = \frac{1}{K}\sum_{k=1}^{K}f_i^k 
\end{equation}
This averaging operation yields a robust and stable representation of the underlying identity, effectively smoothing out transient visual noise and emphasizing consistent appearance cues across time. Within a single camera $c_t$, the collection of all prototypes forms the intra-camera prototype set $\mathcal{P}_{c_t} = \{ p_i \}_{i=1}^{N_{c_t}}$, and the union of prototype sets from all cameras under the same modality constitutes the intra-modality prototype set $\mathcal{\textbf{P}}^m= \{\mathcal{P}_{c_t}^m \}_{t=1}^{C^m}$ ($C^m$ denotes the number of cameras in modality m).
The resulting prototypes serve as high-quality pseudo-centers for subsequent self-supervised or contrastive learning stages.

\begin{algorithm*}[ht]
	\caption{\textbf{:}Training procedure of the proposed HiTPro  } %
	{\bf Input:} 
	Visible and infrared training tracklets  $\mathcal{V}$ and $\mathcal{R}$; \\
	{\bf Initialize:} %
	Temporal-aware Feature Encoder TFE; maximal epoch E; maximal iteration $\mathcal{I}$; 
	\begin{algorithmic}[1]
		\State \textbf{for} e=1 : E \textbf{do} \\
		\hspace*{0.20in} Split  tracklets within each camera into sub-tracklets.\\
		\hspace*{0.20in}  Freeze TFE andconstruct intra-camera prototypes $\mathcal{P}_{c_t}$ by Eq.(\ref{eq:fTFE}) - (\ref{eq:constructp}) and 	 form the modality prototype sets $\mathbf{P}^v, \mathbf{P}^{in}$; \\
		\hspace*{0.20in} \textbf{for} $\mathcal{P}_{c_t}^v$ in $\mathbf{P}^v$ \textbf{do} \\
		\hspace*{0.40in} \textbf{for} $p_i^v$ in $\mathcal{P}_{c_t}^v$  \textbf{do} \\
		\hspace*{0.60in} Calculate current $h_i$  by Eq. (\ref{eq:h1}) - (\ref{eq:h2});\\
		\hspace*{0.60in} Generate cross-cam positive set $J_i^v$ by Eq. (\ref{eq:intra-search1}) - (\ref{eq:intra-search2}) and corresponding weight $\mathcal{W}_i^v$ by Eq. (\ref{eq:w});\\
		\hspace*{0.60in} Generate cross-modality positive set $J_i^{v\to in}$ by Eq. (\ref{eq:cross-search1}) - (\ref{eq:cross-search2}); and corresponding weight  $\mathcal{W}_i^{v\to in}$ by Eq. (\ref{eq:w});\\
		\hspace*{0.40in}\textbf{ end for}\\
		\hspace*{0.20in}\textbf{ end for}\\
		\hspace*{0.20in}  Repeat steps 4 - 10 for $\textbf{P}^{in}$ to obtain $W^{in}$ and  $W^{in \to v}$;\\
		\hspace*{0.20in} \textbf{for} iter=1:$\mathcal{I}$ \textbf{do} \\
		\hspace*{0.40in} Activate  TFE for training and extract batch features $B^v$, $B^{in}$.\\
		\hspace*{0.40in} Calculate $\mathcal{L}_{ic}$ by Eq.(\ref{eq:intra-cam-loss});\\
		\hspace*{0.40in} \textbf{if } $e\ge e_{intra}$ \textbf{then} calculate $\mathcal{L}_{imcc}$ by Eq.(\ref{eq:imcc});\\
		\hspace*{0.40in} \textbf{if } $e\ge e_{cross}$ \textbf{then} calculate $\mathcal{L}_{cm}$ by Eq.(\ref{eq:cm});\\
		\hspace*{0.40in} Calculate $\mathcal{L}_{total}$ and backward to update all prototypes by Eq. (\ref{eq:update});\\
		\hspace*{0.20in} \textbf{end for }\\
		\textbf{end for}
		
	\end{algorithmic}\label{algorithm1}
\end{algorithm*}

\subsection{Hierarchical Cross Prototype Alignment (HCPA)}\label{HCPA}
To bridge representations across cameras and modalities, we propose a Hierarchical Cross Prototype Alignment  strategy that first performs intra-modality cross-camera positive mining, followed by cross-modal extension. 


\textbf{Intra-modality Cross-cam Positive Mining.} For each source prototype ${p}_{i}$ from camera $c_s$,  we compute its similarity to every prototype ${p}_{j}$ in all other cameras $c_t \ne c_s$ of the same modality. For each target camera $c_t$, we select the prototype with the highest similarity to ${p}_{i}$ as the per-camera best candidate:
\begin{equation}\label{eq:intra-search1}
j^*(c_t) = \arg\max_{j \in \mathcal{P}_{c_t}} \mathrm{Sim}({p}_i, {p}_j)
\end{equation}
where $\mathrm{Sim}({p}_i, {p}_j) = \frac{{p}_i^\top {p}_j}{\|{p}_i\| \|{p}_j\|}$. $\mathcal{P}_{c_t}$ denotes the set of prototypes in camera $c_t$. Since the number of times the same pedestrian appears across different cameras is uncertain, we employ a similarity threshold to filter out ambiguous matches. Specifically, for each cross-camera prototype set $\mathcal{P}_{c_t}$, if the prototype  ${p}_{j^*(c_t)}$ most similar to ${p}_i$ exceeds a predefined similarity threshold,  ${p}_{j^*(c_t)}$ is included in the positive sample set.
The final set of reliable intra-modality cross-camera positive indices for $p_i$ is:
\begin{equation}\label{eq:intra-search2}
\mathcal{J}_i = \left\{ j^*(c_t) \mid c_t \neq c_s,\ \mathrm{Sim}({p}_i, {p}_{j^*(c_t)}) \geq h \right\}
\end{equation}
For the visible and infrared modalities, we can obtain $\mathcal{J}_i^v$ and $\mathcal{J}_i^{in}$, respectively.
A detailed discussion on the calculation of the threshold h is provided below.

\textbf{Cross-modality Positive Mining.} While intra-modality cross-camera mining provides high-confidence positive pairs within each modality, bridging the semantic gap between visible and infrared domains requires a dedicated cross-modality alignment stage. We therefore perform cross-modality positive mining separately. Specifically, for a source prototype  ${p}_{i}^v$ from the visible modality, we search for its positive matches among all prototypes in the infrared modality. Then per-cam best infrared candidate is selected as:
\begin{equation}\label{eq:cross-search1}
j^*(c_t) = \arg\max_{j \in \mathcal{P}_{c_t}^{in}} \mathrm{Sim}({p}_i^v, {p}_j^{in})
\end{equation}
where $\mathcal{P}_{c_t}^{in}$ denotes the set of infrared prototypes in infrared camera $c_t^{in}$. Next, we compute reliable cross-modality positive samples using a similar approach:
\begin{equation}\label{eq:cross-search2}
\mathcal{J}_i^{v\to in} = \left\{ j^*(c_t) \mid \mathrm{Sim}({p}_i^v, {p}_{j^*(c_t)}^{in}) \geq h  \right\}
\end{equation}
The positive set  from infrared to visible modality $\mathcal{J}_i^{in\to v}$  can be  derived in the same manner.

\textbf{Dynamic Threshold Strategy (DTS).} While the fixed-threshold  provides a simple baseline, it fails to account for the evolving quality of prototype representations during training. Early in training, features are noisy and inter-identity similarities are compressed, making even true matches fall below a rigid threshold. Conversely, in later stages, over-conservative thresholds may exclude valid positives that exhibit natural appearance variation. To address this, we replace the static threshold with a dynamic, instance-adaptive threshold that scales with the current matching confidence of each query prototype. Specifically, for a source prototype ${p}_{i}$, let $s_{max}^{(i)}$ denote the maximum similarity between  ${p}_{i}$ and its per-camera best matches across all target cameras (within the same modality for intra-modality mining, or across modalities for cross-modality mining). We define the reliability threshold  at epoch e as:
\begin{equation} \label{eq:h1}
h_i = \rho(e) \cdot s_{\max}^{(i)}
\end{equation}
where $\rho(e)$ is a linearly decaying scheduling function:
\begin{equation} \label{eq:h2}
\rho(e) = \rho_{\text{init}} + (\rho_{\text{final}} - \rho_{\text{init}}) \cdot \frac{e}{e_{total}}
\end{equation}
This formulation ensures that the mining process remains sensitive to both the instance-level matching difficulty and the global learning stage and the global learning stage,  yielding a high-quality positive set throughout training.

\textbf{Soft Weight Assignment (SWA).} After identifying the set of reliable positive candidates, we further refine their contribution by assigning soft weights that reflect their relative matching confidence. Specifically, let $\{ s_{ij} = \mathrm{Sim}({p}_i, {p}_j) \mid j \in \mathcal{J}_i \}$ denote the raw similarities of the reliable candidates. We compute normalized weights using a temperature-scaled softmax:

\begin{equation}\label{eq:w}
w_{ij} = \frac{\exp(s_{ij} / \tau_{w})}{\sum_{j' \in \mathcal{J}_i} \exp(s_{ij'} / \tau_{w})}
\end{equation}
where $\tau_{w}$ is a temperature parameter that controls the sharpness of  weight distribution. The final output for prototype  ${p}_i$ is a weighted positive set: $\mathcal{W}_i = \left\{ (j, w_{ij}) \mid j \in \mathcal{J}_i \right\}$. We apply the same computational procedure to $\mathcal{W}_i^{v}$ , $\mathcal{W}_i^{in}$, $\mathcal{W}_i^{v\to in}$, and $\mathcal{W}_i^{in\to v}$.

\subsection{Hierarchical Contrastive Learning (HCL)} \label{HCL}
During training, we perform learning in a hierarchical manner—from intra-camera to intra-modality, and finally to cross-modality. The positive pairs used in training are pre-computed at the beginning of each epoch via the HCPA module described above.  For each visible or infrared training batch $B^{v(in)}$, we randomly select C cameras, and within each camera, we sample P complete tracklets. For each tracklet, we further randomly select S sub-tracklets for training, resulting in a batch-size of $B^{v(in)}=C\times P\times S$. Next, we take a visible  sub-tracklet training feature $q_i^v\in B^{v}$ as an example to illustrate the computation of each loss term. For the infrared modality, the corresponding losses can be derived symmetrically using an analogous approach. Ultimately, each loss term is the sum of the losses computed over  $B^v$ and $B^{in}$.

\textbf{Intra-Camera Loss.}We first calculate intra-camera loss as the basic loss function to encourage consistency among sub-tracklets belonging to the same prototype within a single camera view:

\begin{equation} \label{eq:intra-cam-loss}
\mathcal{L}_{ic}^v = -\frac{1}{B^v}\sum_{{q}_i^v\in B^v}\log \frac{ \exp({q}_i^v ({p}_i^v)^\top / \tau)}{\sum_{j \in \mathcal{C}_i} \exp({q}_i^v({p}_k^v)^\top  / \tau)} 
\end{equation}
where ${p}_i^v$ denotes the prototype feature corresponding to the sub-tracklet feature ${q}_i^v$  and $\mathcal{C}_i$ denotes the set of all prototypes belonging to the same camera as  ${q}_i^v$.
$\tau$ is the temperature parameter.

\textbf{Intra-Modality Cross-Camera Loss.} After the model has learned person identities within each camera, we extend the loss to intra-modality cross-camera alignment, aiming to reduce the distance between ${q}_i^v$ and its cross-camera positive prototypes:
\begin{equation}\label{eq:imcc}
\mathcal{L}_{imcc} = -\frac{1}{B^v}\sum_{{q}_i^v\in B}\sum_{{p}_j^v\in \mathcal{J}_i^{v}} w_{ij} \cdot \log \frac{ \exp({q}_i^v ({p}_j^v)^\top / \tau)}{\sum_{k \in \mathcal{C}_j} \exp({q}_i^v ({p}_k^v)^\top  / \tau)}
\end{equation}
The intra-modality cross-camera loss $\mathcal{L}_{intra-mod}^v$ and $\mathcal{L}_{intra-mod}^{in}$ can be computed using the above formulation.

\textbf{Cross-Modality Loss.} Finally,  the model achieves modality-irrelevant identity alignment by pulling the given query feature  closer to its cross-modality positive prototypes. Given visible sub-tracklet query feature $q_i^v$ and corresponding positive prototypes $\mathcal{J}_i^{v\to in}$, the cross-modality loss is computed by  $\mathcal{L}_{cm}^v=$ 

\begin{equation}\label{eq:cm}
-\frac{1}{B^v}\sum_{{q}_i^v\in B^v}\sum_{{p}_j^{in}\in \mathcal{J}_i^{v\to in}} w_{ij} \cdot \log \frac{ \exp({q}_i^v ({p}_j^{in})^\top / \tau)}{\sum_{k \in \mathcal{C}_j} \exp({q}_i^v({p}_k^{in})^\top  / \tau)}
\end{equation}

\begin{table*}[ht]
	\caption{Comparison with the state-of-the-art methods  on HITSZ-VCM dataset. The best results are highlighted in bold.}
	\label{tab:HITSZ}
	\begin{center}
		\resizebox{\textwidth}{!}{
			\begin{tabular}{c|c|c|c|cccc|cccc}	
				\hline
				& &	& & \multicolumn{4}{c|}{ Infrared to Visible} & \multicolumn{4}{c|}{Visible to Infrared} \\ 				
				\hline
				& Method&Venue &Type	& R1  &  R5 &R10  & mAP  	& R1  &  R5 &R10  & mAP \\		
				\hline
				\multirow{10}{*}{\rotatebox{90}{Supervised}} 		
				&LBA \cite{park2021learning} &ICCV-21 &Image &46.38  &65.29  &72.23   &30.69 &49.30 &69.27 &75.90 &32.38  \\
				&DDAG \cite{Ye2020b} &ECCV-20 &Image &54.62 &69.79 &76.05 &39.26 &59.03 &74.64 &79.53 &41.50  \\
				&CAJ \cite{Ye2021} &ICCV-21  &Image &56.59 &73.49 &79.52 &41.49 &60.13 &74.62 &79.86 &42.81  \\
				
				&MTMML \cite{lin2022learning} &CVPR-22  &Video  &63.74 &76.88 &81.72 &45.31 &64.54 &78.96 &82.98 &47.69 \\
				&IBAN \cite{li2023intermediary} &TCSVT-23 &Video  &65.03 &78.34 &82.98 &48.77 &69.58 &81.51 &85.43 &50.96 \\
				&SADSTRM  \cite{xu2025adversarial} &IJMLC-25 &Video &65.31 &77.86 &82.66 &49.49 &67.66 &80.74 &85.13 &51.76 \\
				&ScRL  \cite{li2025shape} & PR-25 &Video  &71.20 &83.20 &87.40 &52.90 &73.30 &84.40 &87.70 &53.00  \\
				&FA-Net \cite{yang2025fa}  &TIP-25 &Video &68.13 &79.65 &84.01 &51.15  &70.01 &82.11 &85.98 &52.46  \\
				&SIMFGA  \cite{zuo2025spatio} &IVC-25 &Video &75.55 &84.16 &87.33 &64.91 &79.80 &87.98 &90.66 &67.62 \\
				&STHF \cite{tao2026spatial} &TCSVT-26  &Video &70.40 &81.60 &86.20 &56.20 &73.50 &83.70 &87.00 &58.60\\
				\hline
				\multirow{12}{*}{\rotatebox{90}{Unsupervised}} 		
				&C-Contrast \cite{dai2022cluster}  &ACCV-22 &Image  &10.34 &21.43 &27.45 &7.15 &11.18 &23.89 &28.43 &8.21 \\ 
				&ADCA \cite{yang2022augmented} &MM-22 &Image   &15.28 &25.45 &32.69 &13.17 &16.53 &28.29 &36.69 &14.93 \\ 
				&SCA-RCP \cite{li2024inter} &TKDE-24 &Image   &17.47 &29.75 &36.67 &13.21 &18.52 &30.62 &37.32 &13.70 \\
				&PGM \cite{wu2023unsupervised} &CVPR-23 &Image   &19.89 &33.45 &40.64 &17.60 &20.93 &35.30 &41.04 &18.11 \\ 
				&SDCL \cite{yang2024shallow} &CVPR-24 &Image  &20.35 &33.74 &41.90 &19.21 &21.43 &36.74 &41.60 &18.74 \\
				
				&TAUDL \cite{li2018unsupervised} &ECCV-18 &Video &29.10 &45.38 &52.79 &22.64 &30.42 &46.79 &53.28 &23.98  \\
				&UTAL \cite{li2019unsupervised} &TPAMI-19  &Video &33.56 &47.82 &55.02 &25.94&34.67 &48.76 &56.16 &26.85  \\
				&SRC   \cite{xie2022sampling} &TMM-22 &Video &39.93 &56.78 &62.90 &30.79&40.12 &57.01 &63.69 &31.09  \\
				&UFCS   \cite{li2025cross} &IoTJ-25 &Video  &42.98 &58.01 &67.73 &33.90&44.57 &59.26 &68.41 &35.20  \\
				&DRMPCL \cite{zhang2025dual} &Neurocomput-25  &Video &44.48 &59.79 &67.59 &34.23 &45.12 &60.78 &68.20 &37.31\\
				&HiTPro (ours) &- &Video &\textbf{56.58}  &\textbf{70.85} & \textbf{76.08 }&\textbf{44.27} &\textbf{57.98} &\textbf{71.28} &\textbf{77.39} &\textbf{45.70}  \\
				\hline
			\end{tabular}
		}
	\end{center}	
\end{table*}

\textbf{Hierarchical Loss Scheduling (HLS).} To stabilize training and avoid premature exposure to noisy cross-cam or cross-modality matches, we adopt a hierarchical loss scheduling strategy that progressively introduces supervision signals of increasing difficulty. The total loss at epoch e is formulated by:
\begin{equation}
\begin{aligned} \label{eq:total-loss}
\mathcal{L}_{total}(e)&=\mathcal{L}_{ic}+\mathbb{I}[e\ge e_{intra}]\cdot \mathcal{L}_{imcc}+ \mathbb{I}[e\ge e_{cross}]\cdot \mathcal{L}_{cm}
\end{aligned}
\end{equation}
where $\mathbb{I}(\cdot)$ denotes the indicator function, equal to 1 if the condition is true and 0 otherwise. $e_{intra}$ and $e_{cross}$ denote the epoch at which the computation of $\mathcal{L}_{intra-mod}$ and $\mathcal{L}_{cross-mod}$ begin, respectively.

\textbf{Prototype Update Strategy.} Inspired by memory-bank based methods such as Cluster-Contrast\cite{dai2022cluster}, we update the tracklet prototypes using the sub-tracklet features during the back-propagation process. Each sub-tracklet feature  contributes not only to its own prototype but also to all reliably associated positive prototypes identified by the HCPA module. Specifically, given a visible sub-tracklet feature ${q}_i^v$, we update all these prototypes via exponential moving average:

\begin{equation} \label{eq:update}
\begin{aligned}
&{p}_i^v \leftarrow (1-\alpha)\cdot {p}_i^v+\alpha\cdot {q}_i^v\\
&{p}_j^v \leftarrow (1-\alpha)\cdot {p}_j^v+\alpha\cdot {q}_i^v, \forall j \in \mathcal{J}_i^{v}  \\
&{p}_j^{in}\leftarrow (1-\alpha)\cdot {p}_j^{in}+\alpha\cdot {q}_i^v, \forall j \in \mathcal{J}_i^{v\to {in}} 
\end{aligned}
\end{equation}
where $\alpha$ is the momentum coefficient. This strategy keeps prototypes aligned with the evolving feature space of the encoder and actively mitigates cross-camera and cross-modal discrepancies. The overall training algorithm proceeds as Algorithm \ref{algorithm1}.

\section{Experiments}
\subsection{Datasets and Evaluation Metric}
\textbf{HITSZ-VCM\cite{lin2022learning}} is the first large-scale dataset that jointly incorporates video tracklets and dual visible-infrared modalities. The dataset was collected using 12  visible and 12 corresponding infrared cameras, capturing pedestrian trajectories across diverse indoor and outdoor scenes including  passageways, playgrounds, and gardens. It contains 927 unique identities, with a total of 463,259 frames organized into 21,863 tracklets by default (11,785 in the visible modality and 10,078 in the infrared modality). The original author split the training set into  11,061 video sequences corresponding to 500 distinct identities, whereas the test set consists 10,802 video sequences from 427 identities.In this work, for the purpose of unsupervised learning, we do not adopt the original authors' training set partitioning strategy for tracklet segmentation. Instead, we preserve the continuous trajectory of the same pedestrian within each camera while discarding their identity labels to fulfill the objective of unsupervised learning, resulting in a training set of 1481 visible tracklets and 1445 infrared tracklets of 500 identities. The test set remains unchanged. 

\textbf{BUPTCampus\cite{du2023video}} is a large-scale, video-based visible-infrared person re-identification dataset which is collected using 6 binocular dual-modality cameras that simultaneously record RGB and infrared  images of individuals. The dataset contains 3,080 unique identities, totaling 1,869,066 images and is organized into 16,826 video sequences.  Following the same protocol as HITSZ-VCM, we retain complete tracklets in the training set, resulting in a total of 2,335 visible and 2,335 infrared trajectories, covering 1,000 distinct identities. This work focuses on unsupervised video person re-identification and therefore does not utilize the auxiliary annotations provided in this dataset.

The  Cumulative Matching Characteristics (CMC) and mean Average Precision (mAP) are employed as evaluation metrics.

\subsection{Implementation Details}
The experiments are conducted using a single 24G RTX 4090 GPU and the algorithm are based on Pytorch. SGD is employed as optimizer with lr=0.00035, momentum=0.9 and learning rate is decayed to 0.1 of its initial value every 20 epochs. The total training epochs $e_{total}$ is 60, and for each epoch the training iters $\mathcal{I}$ is 300.  For both training and feature extraction, we adhere to the original dataset configurations: we set seq\_len=6 for HITSZ-VCM and seq\_len=10 for BUPTCampus.
We resized each image to $288\times 144 $ and data augmentation methods such as Padding, RandomCrop, RandomHorizontalFlip, ChannelRandomErasing and ChannelExchange\cite{Ye2021} are employed for training images. 
For ICTP module, the number of sub-tracklets K is set to 4. For HCPA module,  $\tau_s$ and $\tau_{w}$ is set to 0.1. $\rho_{init}$ and $\rho_{final}$ is set to 0.99 and 0.90, respectively.  During HCL training stage, each batch  contains both visible and infrared modalities with  2 cameras $\times$ 2 full tracklet $\times$ 2 sub-tracklets (which means $B^{(v(in)}=2\times2\times2=8$). Temperature $\tau=0.05$. Momentum $\alpha=0.2$. $e_{intra}$ and $e_{cross}$ is set to 5, 15, respectively.

\begin{table*}[htbp]
\caption{Comparison with the state-of-the-art methods  on BUPTCampus  dataset. The best results are highlighted in bold.}
\label{tab:BUPT}
\begin{center}
\resizebox{\textwidth}{!}{
	\begin{tabular}{c|c|c|c|cccc|cccc}	
		\hline
		& &	& & \multicolumn{4}{c|}{ Infrared to Visible} & \multicolumn{4}{c}{Visible to Infrared} \\ 				
		\hline
		& Method&Venue &Type	& R1  &  R5 &R10  & mAP  	& R1  &  R5 &R10  & mAP \\		
		\hline
		\multirow{6}{*}{\rotatebox{90}{Supervised}} 		
		&LBA \cite{park2021learning} &ICCV-21 &Image &32.09 &54.85 &65.11 &32.93 &39.07 &58.70 &66.48 &37.06 \\
		&CAJ \cite{Ye2021} &ICCV-21  &Image &40.49 &66.79 &73.32 &41.46 &45.00 &70.00 &77.04 &43.61  \\
		&DEEN \cite{zhang2023diverse} &CVPR-23  &Image & 49.81 &71.64 &80.97 &48.59 &53.70 &74.81 &80.74 &50.43\\ 
		&MTMML \cite{lin2022learning} &CVPR-22  &Video  &40.64 &57.66 &63.52 &38.13 &42.97 &59.57 &66.02 &42.97 \\
		&SADSTRM  \cite{xu2025adversarial} &IJMLC-25 &Video &54.21 &72.03 &80.27  &51.65 &53.12 &74.80 &81.54  &49.65\\
		&SIMFGA  \cite{zuo2025spatio} &IVC-25 &Video  &55.05 &73.75 &80.46 &54.57 &54.88 &76.56 &82.62 &51.76  \\
		\hline
		\multirow{9}{*}{\rotatebox{90}{Unsupervised}} 		 
		&ADCA \cite{yang2022augmented} &MM-22 &Image   &1.87 &6.90 &11.19 &4.39 &2.96 &7.22 &11.11 &4.84 \\ 
		&SCA-RCP \cite{li2024inter} &TKDE-24 &Image   &2.43 &5.60 &8.96 &3.43 &3.89 &8.52 &14.26 &5.41 \\
		&PGM \cite{wu2023unsupervised} &CVPR-23 &Image   &4.10 &8.77 &14.74 &5.51 &5.37 &13.52 &19.63 &8.16 \\ 
		&SDCL \cite{yang2024shallow} &CVPR-24 &Image  &4.91 &9.22 &15.68 &6.02 &5.87 &16.71 &23.45 &9.56 \\
		&TAUDL \cite{li2018unsupervised} &ECCV-18 &Video &18.84 &32.46 &39.55 &19.76 &17.41 &34.44 &40.37 &19.93  \\
		&UTAL \cite{li2019unsupervised} &TPAMI-19  &Video  &19.03 &33.40 &39.74 &22.64 &30.42 &46.79 &53.28 &23.98   \\
		&UFCS   \cite{li2025cross} &IoTJ-25 &Video  &21.46 &38.81 &46.27 &23.27 &23.70 &38.52 &46.30 &24.10  \\
		&DRMPCL \cite{zhang2025dual} &Neurocomput-25  &Video &23.88 &40.30 &46.64 &25.12 &24.63 &40.74 &48.89 &25.51\\
		&HiTPro (ours) &- &Video &\textbf{42.91}  &\textbf{62.13} & \textbf{69.22}&\textbf{41.78} &\textbf{ 42.22} &\textbf{65.37} &\textbf{71.30} &\textbf{41.05}  \\
		\hline
	\end{tabular}
}
\end{center}	
\end{table*}

\subsection{Comparison with the state-of-the-art methods}
We conduct a comprehensive comparison between our proposed method and current state-of-the-art (SOTA) approaches on  HITSZ-VCM  and BUPTCampus datasets, including  supervised methods  and unsupervised methods.
These methods  can be generally categorized into image-based methods and video-based methods. For cross-modal, image-level unsupervised methods such as PGM \cite{wu2023unsupervised} , we adapt them to video dataset by replacing image features with averaged video frame features. Since no existing unsupervised cross-modal video-level learning methods are available, we adapt several single-modal unsupervised video learning approaches , such as UTAL \cite{li2019unsupervised}, UFCS   \cite{li2025cross}, to the cross-modal setting.

The experimental results on  HITSZ-VCM  are shown in Table. \ref{tab:HITSZ}. It is obvious that The proposed HiTPro method in this work significantly outperforms other unsupervised learning approaches and  achieves performance that is comparable to or even surpasses that of several supervised learning methods. Specifically,under infrared to visible mode, our method outperforms the second-best unsupervised approach by 12.10\% in Rank-1 accuracy and 8.04\% in mAP.  For image-based unsupervised methods, the lack of video-level feature modeling capability—coupled with their reliance on pseudo-labels generated by image clustering algorithms—leads to sub-optimal performance on unsupervised cross-modal video datasets. Video-based unsupervised learning methods, such as UFCS \cite{li2025cross}, DRMPCL \cite{zhang2025dual}, achieve modest improvements in accuracy. However, they still underperform compared to our method when applied to cross-modal datasets. Our method achieves consistently strong performance under both visible-to-infrared and infrared-to-visible evaluation settings, maintaining state-of-the-art results. The results confirm that the proposed HiTPro learning framework achieves significant improvements over existing solutions, particularly in the demanding unsupervised  video-based cross-modal setting.

We also conduct experiments on the newly proposed BUPTCampus dataset, with results reported in Table \ref{tab:BUPT}. Our method also achieves the best performance  under unsupervised mode, reporting 42.91\% Rank-1 accuracy and 41.78\% mAP under infrared to visible mode,  outperforming the second-best  approach by 19.03\% in Rank-1 accuracy and 16.66\% in mAP. It even surpass several supervised image-based methods in the process. The absolute performance on BUPTCampus is moderately lower than on HITSZ-VCM, primarily due to its larger cross-modality gap and more dynamic scene variations. Despite these challenges, HiTPro maintains a clear margin over all baselines, demonstrating its robustness in realistic and demanding environments. Overall, these results underscore that our proposed framework generalizes well beyond a single benchmark and sets a new state of the art for unsupervised video-based visible-infrared person re-identification.

\subsection{Ablation study}

\begin{table*}[ht]
\caption{Effectiveness of the each components of the proposed HiTPro on the HITSZ-VCM dataset.}
\label{tab:abl}
\begin{center}\resizebox{0.9\textwidth}{!}{
	\begin{tabular}{c|cccc|cc|cc}
		\hline
		&\multicolumn{4}{c|}{Components} & \multicolumn{2}{c|}{Infrared to Visible}    & \multicolumn{2}{c}{Visible to Infrared}\\
		\hline 
		index &Baseline &TFE &ICTP &HCPA\&HCL   &R1 &mAP &R1 &mAP  \\
		\hline 
		1&$\checkmark$ & & &  &10.73 &6.70  &11.24 &7.05  \\
		2&$\checkmark$ &$\checkmark$ & &  &19.77 &15.16  &21.45 &17.91  \\
		3&$\checkmark$ & &$\checkmark$ &  &20.64 &15.89  &22.09 &18.83  \\
		4&$\checkmark$ &$\checkmark$ &$\checkmark$  & &31.62 &21.55  &33.02 &22.89   \\		
		5&$\checkmark$&$\checkmark$&$\checkmark$ &$\checkmark$ &\textbf{56.58} &\textbf{44.27} &\textbf{57.98} &\textbf{45.70}  \\
		\hline
	\end{tabular}
}
\end{center}
\end{table*}
In this section, we conduct ablation studies on the HITSZ-VCM dataset to evaluate the individual effectiveness of each component in the proposed method. The results are shown in Table. \ref{tab:abl}. The baseline model employs a standard AGW \cite{ye2021deep} backbone based on ResNet-50 without temporal modeling. All other settings in these experiments are kept consistent with the optimal configuration. During training, it also incorporates the  Cluster-Contrast \cite{dai2022cluster} framework, a clustering-based unsupervised ReID learning paradigm. As shown in the first row of the table, conventional unsupervised approaches based on image-level features and clustering perform poorly on cross-modal video ReID, achieving only a Rank-1 accuracy of 10.73\% under infrared to visible mode.

\textbf{Effectiveness of TFE.} We first investigate the contribution of the Temporal-aware Feature Encoder (TFE) to the overall model. By comparing experiments index 1 vs. 2 and 3 vs. 4 in Table \ref{tab:abl}, we observe that the TFE module significantly improves the model’s video re-identification accuracy over the baseline. We further investigate the role and necessity of the individual components within the TFE, with results presented in Table \ref{tab:TFE}.  Incorporating the Transformer-based Temporal Encoder (TTE)  yields a substantial gain, boosting Rank-1 to 50.10\% and mAP to 38.13\% under infrared to visible mode, as shown in index 2. This confirms that modeling long-range dependencies across frames effectively captures motion patterns and occlusion dynamics, which are crucial for video-based Re-ID. Besides, adding the Adaptive Frame-weighting Module (AFM) alone also yields a notable improvement, increasing Rank-1 accuracy by 8.68\% compared to the baseline (as show in index 3). This component allows the network to dynamically emphasize informative frames while suppressing noisy or occluded ones, leading to more robust and discriminative video-level embeddings. The combination of these modules achieves the best overall performance (as shown in index 4), demonstrating the synergistic effectiveness of the proposed modules in extracting discriminative features for video-based person re-identification.

\begin{table}[ht]
\caption{Effectiveness of the each components of the Temporal-aware Feature Encoder (TFE) on the HITSZ-VCM dataset.}
\label{tab:TFE}
\begin{center}\resizebox{0.5\textwidth}{!}{
	\begin{tabular}{c|ccc|cc|cc}
		\hline
		&\multicolumn{3}{c|}{Components} & \multicolumn{2}{c|}{Infrared to Visible}    & \multicolumn{2}{c}{Visible to Infrared}\\
		\hline 
		index &Baseline &TTE &AFM &R1 &mAP &R1 &mAP  \\
		\hline 
		1&$\checkmark$ &  & &40.13 &31.28  &49.36 &30.02  \\
		2&$\checkmark$ &$\checkmark$ &  &50.10 &38.13  &51.25 &39.16 \\
		3&$\checkmark$ & &$\checkmark$ &48.81 &37.61  &49.13 &36.54  \\
		4&$\checkmark$&$\checkmark$&$\checkmark$ &\textbf{56.58} &\textbf{44.27} &\textbf{57.98} &\textbf{45.70}  \\
		\hline
	\end{tabular}
}
\end{center}
\end{table}

\begin{table*}[ht]
	\caption{The effectiveness of the components in  HCPA  process and the individual loss terms in the HCL training stage on the HITSZ-VCM dataset. 'DTS' denotes Dynamic Threshold Strategy. 'SWA' denotes Soft Weight Assignment. 'HLS' denotes Hierarchical Loss Scheduling.}
	\label{tab:HCPA}
	\begin{center}\resizebox{0.9\textwidth}{!}{
			\begin{tabular}{c|cccccc|cc|cc}
				\hline
				&\multicolumn{6}{c|}{Components} & \multicolumn{2}{c|}{Infrared to Visible}    & \multicolumn{2}{c}{Visible to Infrared}\\
				\hline 
				index &$\mathcal{L}_{ic}$ &$\mathcal{L}_{imcc}$ &$\mathcal{L}_{cm}$  &HLS    &DTS &SWA &R1 &mAP &R1 &mAP  \\
				\hline 
				1&$\checkmark$ & &  & & &   &31.62 &21.55  &33.02 &22.89  \\
				2&$\checkmark$ &$\checkmark$ & &  & & &40.01 &34.44  &45.89 &35.46  \\
				3&$\checkmark$ & &$\checkmark$ &  & & &15.05 &10.77  &15.64 &11.03  \\
				4&$\checkmark$ &$\checkmark$ &$\checkmark$&  & &  &45.61 &38.33  &45.92 &36.87    \\
				5&$\checkmark$ &$\checkmark$ &$\checkmark$ &$\checkmark$ &  & &50.96 &39.58 &51.21 &40.14 \\ 
				6&$\checkmark$ &$\checkmark$ &$\checkmark$ &$\checkmark$ &$\checkmark$ & &54.38 &42.18 &55.12 &42.36     \\			
				7&$\checkmark$&$\checkmark$&$\checkmark$ &$\checkmark$  &$\checkmark$ &$\checkmark$ &\textbf{56.58} &\textbf{44.27} &\textbf{57.98} &\textbf{45.70}  \\
				\hline
			\end{tabular}
		}
	\end{center}
\end{table*}

\begin{figure}[hbpt]
	\centering
	\includegraphics[width=1\linewidth]{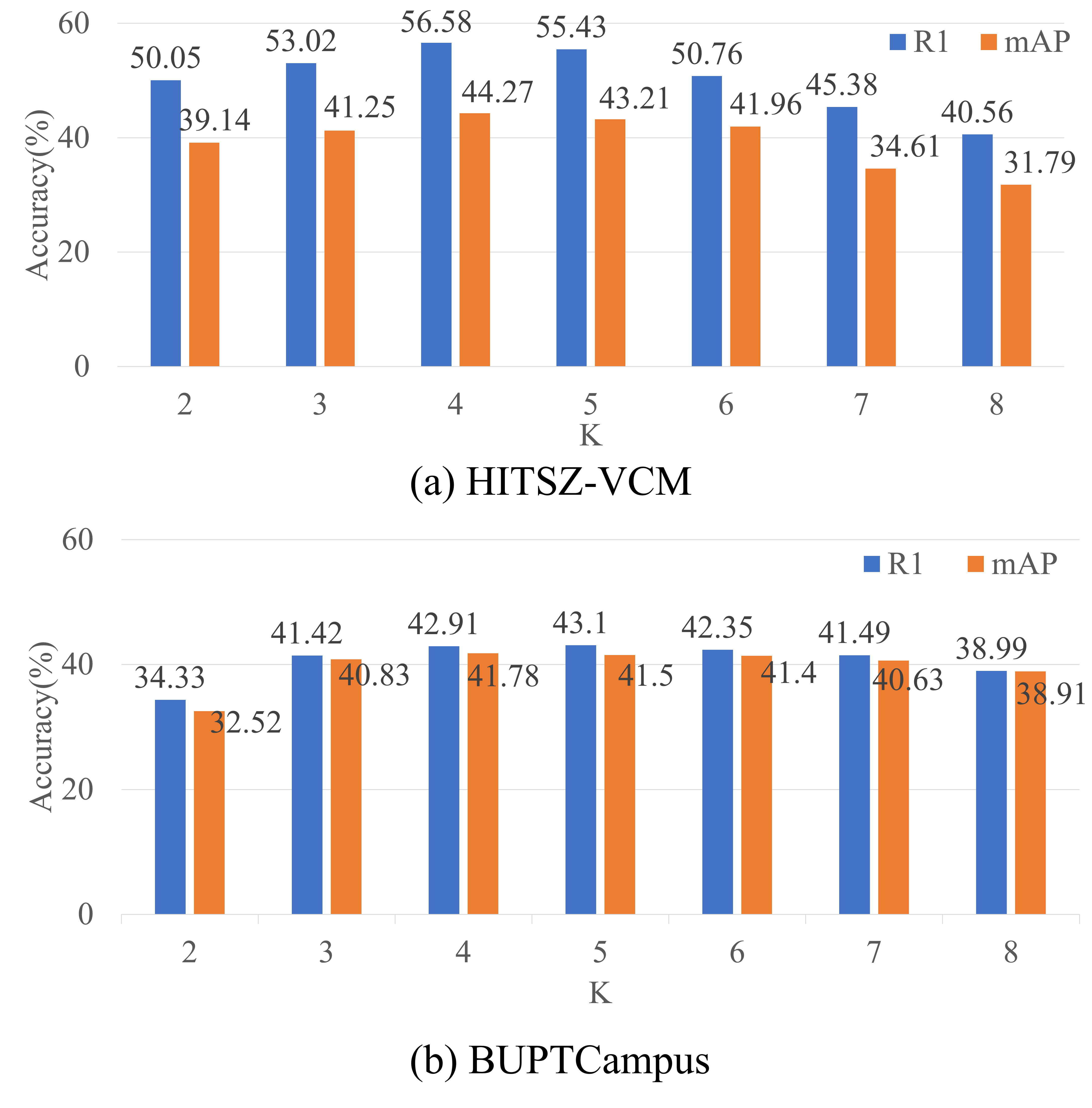}
	\caption{Influence of K on (a) HITSZ-VCM dataset and (b) BUPTCAMPUS dataset under infrared to visible test mode. Rank-1 and mAP are reported. }
	\label{fig:K}
\end{figure}

\begin{figure}[ht]
	\centering
	\includegraphics[width=1\linewidth]{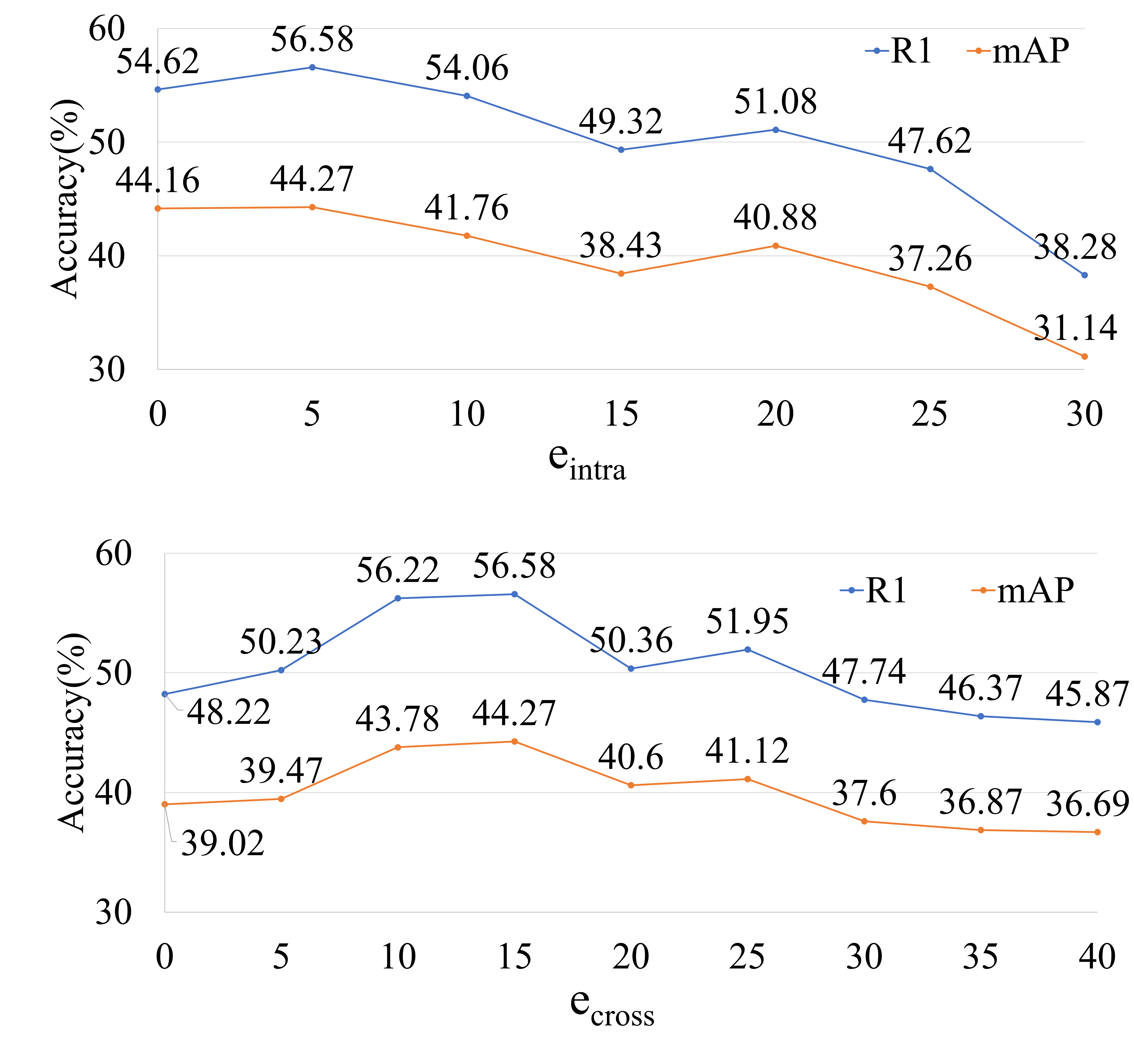}
	\caption{Influence of K on (a) HITSZ-VCM dataset and (b) BUPTCAMPUS dataset under infrared to visible test mode. Rank-1 and mAP are reported. }
	\label{fig:e-epoch}
\end{figure}

\begin{figure}[ht]
	\centering
	\includegraphics[width=1\linewidth]{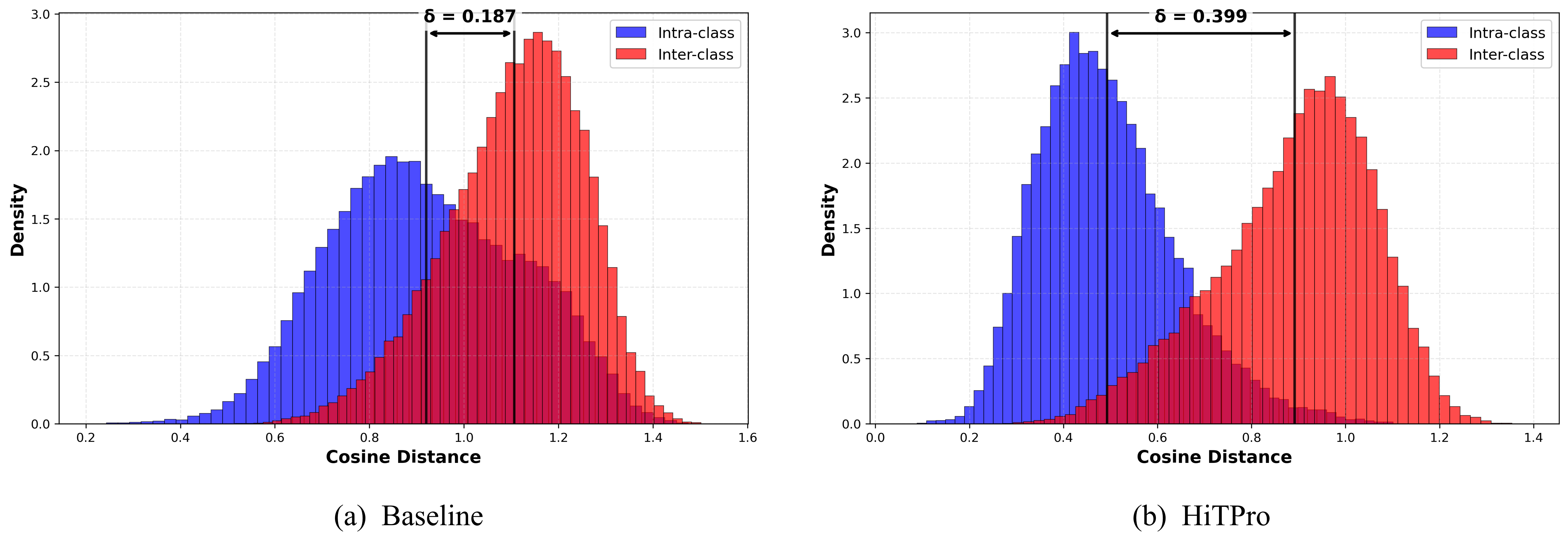}
	\caption{Cosine distance distributions of 40,000 randomly sampled positive (intra-class, blue) and negative (inter-class, red) pairs on the test set. Left: baseline model. Right: our HiTPro. }
	\label{fig:dist}
\end{figure}

\begin{figure*}[ht]
	\centering
	\includegraphics[width=0.8\linewidth]{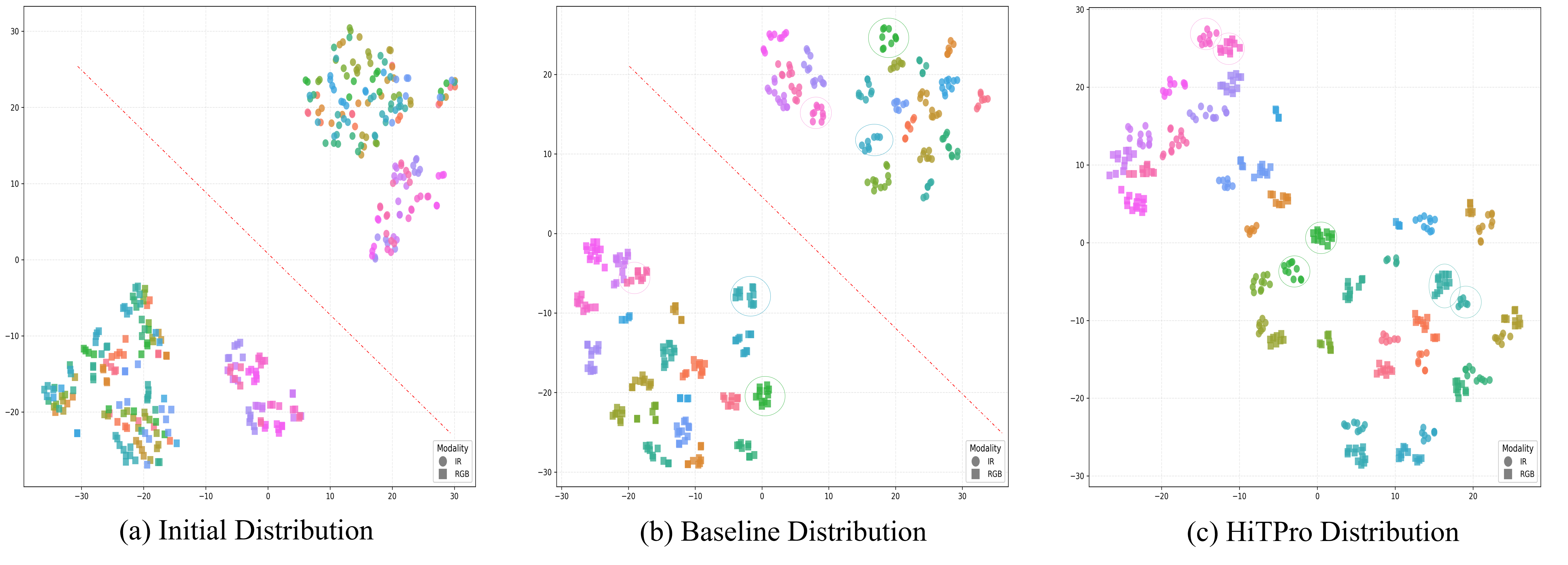}
	\caption{t-SNE visualization of tracklet features for 20 randomly selected identities from the test set. Each color represents a unique identity; circles denote infrared features and squares denote visible features. (a) Initial embedding before training. (b) Features learned by  baseline  method. (c) Features learned by our HiTPro, achieving compact and modality-invariant identity clusters. }
	\label{fig:tsne}
\end{figure*}

\begin{figure*}[ht]
	\centering
	\includegraphics[width=0.8\linewidth]{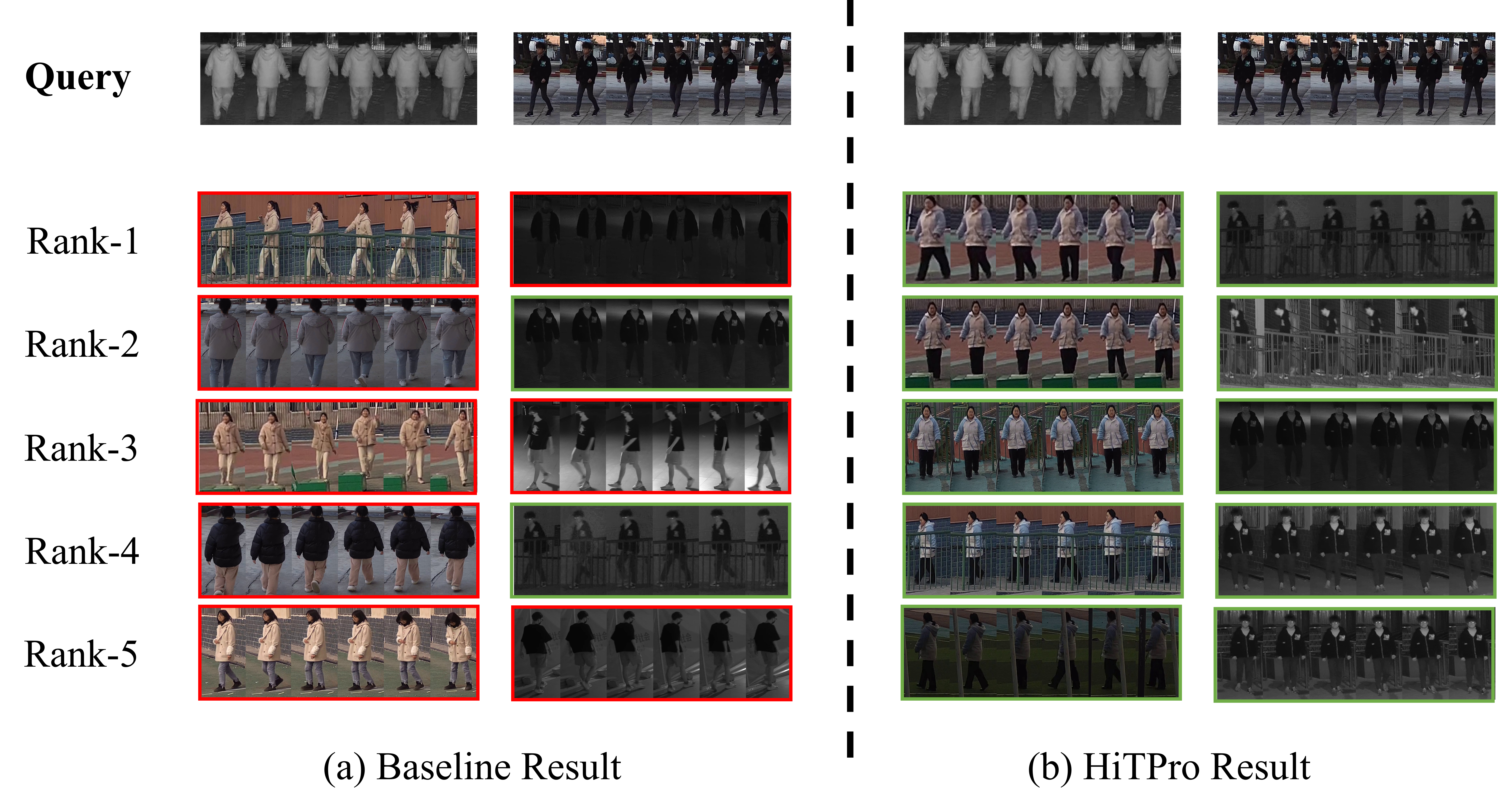}
	\caption{Top-5 Retrieval results of (a) Baseline and (b) HiTPro method on HITSZ-VCM dataset. Correct matches are highlighted with green boxes, while incorrect retrievals are marked in red boxes.}
	\label{fig:retrieval}
\end{figure*}

\textbf{Effectiveness of ICTP.} Next, we evaluate the contribution of the proposed Intra-Camera Temporal Prototyping (ICTP) module.  The baseline model processes video features using conventional clustering methods, while ICTP module construct trajectory-level prototypes by averaging features from sub-tracklets within a single camera view, thereby providing reliable pseudo-identities for self-supervised learning.  Adding the ICTP module alone means that only the Intra-Camera loss $\mathcal{L}_{ic}$ is incorporated during training. By comparing index 3 vs. 1 and index 4 vs. 2 in Table \ref{tab:abl}, we observe that the introduction of ICTP delivers a noticeable performance boost—incorporating  ICTP alone brings approximately a 10\% gain in Rank-1 accuracy. This  improvement demonstrates that ICTP effectively leverages the identity-disjoint property of intra-camera tracklets to form stable and discriminative prototype representations. More importantly, ICTP serves as a foundational component for the subsequent HCPA matching process and the HCL training paradigm. More importantly, ICTP serves as a foundational component for the subsequent HCPA  process and the HCL training paradigm.

\textbf{Effectiveness of HCPA and HCL.} We finally evaluate the effectiveness of the proposed  Hierarchical Cross Prototype Alignment (HCPA) and Hierarchical Contrastive Learning (HCL) modules. Due to the strong coupling and high synergy between the two modules, we evaluate their performance together as an integrated component. By  Comparing index 4 and 5  in Table \ref{tab:abl}, after incorporating HCPA and HCL, the model achieves a significant performance gain: Rank-1 accuracy improves by 24.96\% and mAP increases by 22.72\% under infrared to visible mode compared to using only the ICTP learning scheme. This clearly demonstrates that unsupervised contrastive learning—from cross-camera to cross-modal alignment—is key to the model’s performance improvement. To further evaluate the effectiveness of individual components in HCPA and the synergistic contribution of the loss terms in HCL, we conduct more fine-grained ablation studies on the HITSZ-VCM dataset, the results are shown in Table \ref{tab:HCPA}. By comparing index 2 vs. 1, it can be observed that incorporating the Intra-Modality Cross-Camera Loss $\mathcal{L}_{imcc}$ leads to a significant performance gain compared to performing contrastive learning only within individual cameras, demonstrating the importance of intra-modality cross-camera learning. However, as shown by comparing index 3 vs. 1, directly incorporating the cross-modality loss $\mathcal{L}_{cm}$ alone actually degrades model performance. This is likely because the modality gap is significantly larger than the inter-camera variation; thus, introducing cross-modal learning without intra-modality guidance in training introduces substantial noise, leading to a sharp decline in the precision of positive samples.   Comparing index 4 vs. 3, the model achieves a significant performance gain after jointly incorporating the $\mathcal{L}_{imcc}$ and $\mathcal{L}_{cm}$ together, with Rank-1 accuracy improving by  30.56\%. This demonstrates that intra-modal alignment is an indispensable step for effective cross-modal identity alignment.
Incorporating Hierarchical Loss Scheduling (HLS) further improves the model accuracy, as shown in index 5. 
The introduction of Dynamic Threshold Strategy (DTS) enhances the reliability of positive samples during early training and gradually incorporates hard positives in later stages to improve model performance while mitigating noise from falsely labeled positives. Meanwhile, the  Soft Weight Assignment (SWA) module further assigns higher weights to more reliable positive samples. As shown in Index 6 and 7, both modules contribute  Rank-1 accuracy gains of approximately 2–3\%. Together, they validate that explicitly modeling the uncertainty of unsupervised matches, through adaptive mining and confidence-aware weighting, is essential for robust cross-modality alignment.

\subsection{Parameter Analysis}

\textbf{Influence of  sub-tracklets number K.} To further investigate the optimal value of number of sub-tracklets K proposed in ICTP, we conduct  parameter ablation study on both  the HITSZ-VCM and BUPTCAMPUS dataset. As shown in Figure \ref{fig:K}, we evaluate the performance of our HiTPro framework on both HITSZ-VCM and BUPTCampus datasets under varying values of 
K from 2 to 8. On HITSZ-VCM, performance initially increases with K, peaking at K=4, then gradually declines as K increases further. This suggests that partitioning a tracklet into too few segments may fail to capture fine-grained temporal variations, while overly fine segmentation introduces noise and reduces feature stability due to shorter sub-tracklets. The optimal K=4 strikes a balance between temporal granularity and representation robustness. On BUPTCampus, the trend is similar: performance improves from K=2 to K=5, before dropping slightly for larger. The consistent peak around K=4 or K=5 across both datasets indicates that moderate temporal partitioning is crucial for effective prototype learning. These results demonstrate that the ICTP module is sensitive to K , but remains stable within a reasonable range. We therefore set K=4 as the default in all experiments, which provides the best trade-off between temporal modeling fidelity and feature consistency.

\textbf{Influence of $e_{intra}$ and $e_{cross}$.} To evaluate the effectiveness of our hierarchical training strategy, we conduct ablation experiments on the HITSZ-VCM dataset to analyze the impact of the starting epochs for intra-modality cross-camera learning ($e_{intra}$) and cross-modal learning ($e_{cross}$). Specifically, we vary the initiation epoch of these two phases and measure the final performance. As shown in Figure \ref{fig:e-epoch}, the model achieves its highest accuracy when the $e_{intra}$  is set to 5, which suggests that intra-modality cross-camera learning can start at an earlier stage, as the inter-camera discrepancies are relatively small. However, initiating cross-modality alignment too early (before epoch 10) leads to a noticeable drop in accuracy due to unaligned intra-modality representations. Delayed cross-modality learning  limits the model’s ability to fully exploit cross-spectrum information, causing a gradual decline. We therefore select $e_{cross}=15$ as the optimal epoch to start cross-modality learning.

\subsection{Visualization}
\textbf{Feature Distance Distribution.} To  validate the effectiveness of our proposed HiTPro framework in learning discriminative and modality-invariant representations, we conduct a visualization analysis on the  distance distribution between feature pairs. Specifically, we randomly sample 40,000 positive  and negative pairs from the test set and compute their cosine distances. The resulting density distributions are illustrated in Figure \ref{fig:dist}. The baseline model exhibits substantial overlap between intra-class (blue) and inter-class (red) cosine distances, with a small mean gap, indicating weak discriminability. In contrast, after training with our method, the distributions become well-separated. The improved class separability directly supports our method's superior retrieval performance and confirms that HiTPro effectively learns a compact, identity-preserving embedding space without relying on pseudo-labels.

\textbf{TSNE Distribution.} To further illustrate the evolution of feature representations, we visualize the embedding space using t-SNE on 20 randomly selected identities from the test set. In the plot, each color denotes a unique identity, while circle markers represent infrared  tracklet feature and squares  denote visible (RGB) tracklet feature. As shown in Figure \ref{fig:tsne}, (a) the initial distribution reveals no clear identity clustering, with features scattered across the space regardless of modality. In (b), the baseline model, which employing conventional clustering-based contrastive learning, achieves partial intra-modal aggregation, where same-identity instances within RGB or IR are grouped to some extent. However, a significant cross-modality gap remains, as corresponding identities from different modalities are still separated, indicating limited alignment capability. In contrast, (c) our HiTPro framework successfully bridges this gap: features of the same person are tightly clustered across both modalities, forming compact and well-separated identity manifolds. This demonstrates that HiTPro effectively learns modality-invariant representations by aligning sub-tracklet features with identity prototypes in a hierarchical and unsupervised manner.

\textbf{Retrieval Results Analysis}
We present qualitative retrieval results on the test set by visualizing 6  frames sampled from each query tracklet of HITSZ-VCM dataset. As shown in Figure~\ref{fig:retrieval},   we present the Rank-5 results under dual-modal retrieval, where correct matches are highlighted with green boxes and incorrect matches with red boxes.
The baseline frequently suffers from retrieval failures caused by similar appearances, poses, or excessive modality discrepancies. These errors reveal its vulnerability to superficial cues and insufficient cross-modal alignment. In contrast, HiTPro  retrieves the correct identity across under both search mode, even in challenging cases such as low-texture IR imaging, or drastic viewpoint shifts.  The results confirm that HiTPro effectively disentangles identity-relevant features from modality- and pose-dependent noise, leading to more reliable video-based cross-modality retrieval.

\section{Conclusion} 
In this work, we formulate and address the problem of unsupervised video-based visible-infrared person re-identification, a practical yet underexplored task that requires joint modeling of cross-modality gaps and spatio-temporal dynamics in unlabeled videos.
To tackle this problem, we propose HiTPro, a clustering-free framework without explicit pseudo-label assignment that leverages video structure through a efficient Temporal-aware Feature Encoder (TFE) to obtain tracklet representations. It then employs Intra-Camera Tracklet Prototyping (ICTP) for identity exclusivity, followed by Hierarchical Cross-Prototype Alignment (HCPA) with dynamic thresholding and soft weighting to reliably mine cross-modal positives. Finally, Hierarchical Contrastive Learning (HCL) enables progressive optimization from intra-camera to cross-modality alignment.
Extensive experiments on HITSZ-VCM and BUPTCampus show that HiTPro achieves state-of-the-art performance among unsupervised methods.  Ablation studies further validate the synergistic effectiveness of the proposed modules in our framework. 
Nevertheless, a gap remains compared to supervised approaches, and performance may degrade under poor tracking or severe occlusion.
We hope HiTPro serves as a strong baseline for unsupervised video-based VI-ReID and inspires future work on robust prototype-driven learning in cross-modal retrieval, such as handling imperfect tracklets or extending the paradigm to other modalities like text-to-image or audio-visual tasks.


\bibliography{bibforUVVReID}
\bibliographystyle{IEEEtran}
%



%

%

\end{document}